\documentclass[10pt, a4paper]{article}
\usepackage{lrec2022} 
\usepackage{multibib}
\usepackage{graphicx}
\usepackage{tabularx}
\usepackage{soul}
\usepackage{titlesec}
\titleformat{\section}{\normalfont\large\bfseries\center}{\thesection.}{1em}{}
\titleformat{\subsection}{\normalfont\SmallTitleFont\bfseries\raggedright}{\thesubsection.}{1em}{}
\titleformat{\subsubsection}{\normalfont\normalsize\bfseries\raggedright}{\thesubsubsection.}{1em}{}
\renewcommand\thesection{\arabic{section}}
\renewcommand\thesubsection{\thesection.\arabic{subsection}}
\renewcommand\thesubsubsection{\thesubsection.\arabic{subsubsection}}

\usepackage{epstopdf}
\usepackage[utf8]{inputenc}

\usepackage{hyperref}
\usepackage{xstring}
\usepackage[shortlabels]{enumitem}
\usepackage{color}
\usepackage{amsmath}

\def\fj#1{\textcolor{black}{#1}} 

\title{ \textbf{Exploring the GLIDE model for Human Action-effect Prediction}}

\name{Fangjun Li$^1$, David C. Hogg$^1$, Anthony G. Cohn$^{1,2,3,4,5}$  }

\address{
$^1$School of Computing, University of Leeds, UK\\
$^2$Luzhong Institute of Safety, Qingdao University of Science and Technology, China\\
$^3$College of Electronic and Information Engineering, Tongji University, China\\
$^4$School of Mechanical and Electrical Engineering, Qingdao University of Science and Technology, China \\
$^5$School of Control Science and Engineering, Shandong University, China \\
\{scfli, D.C.Hogg, A.G.Cohn\}@leeds.ac.uk\\}

\abstract{
We address the following action-effect prediction task.  Given an image depicting an initial state of the world and an action expressed in text, predict an image depicting the state of the world following the action. The prediction should have the same scene context as the input image. 
We explore the use of the recently proposed GLIDE model for performing this task. GLIDE is a generative neural network that can synthesize (inpaint) masked areas of an image, conditioned on a short piece of text. Our idea is to mask-out a region of the input image where the effect of the action is expected to occur. GLIDE is then used to inpaint the masked region conditioned on the required action. In this way, the resulting image has the same background context as the input image, updated to show the effect of the action. We give qualitative results from experiments using the EPIC dataset of ego-centric videos labelled with actions. 
 \\ \newline  }

\begin{document}

\maketitleabstract

\section{Introduction}

The purpose of this study is to investigate the potential of a generative model to reason about human actions occurring in a complex physical environment. The model will be given a textual description for an action and an initial world state depicted in an image; it needs to predict an image depicting the final world state following the action. E.g., given an initial image depicting someone holding a carrot and a knife, and the action `peel carrot', the model should predict an image in which 'peelings' have been separated from the carrot.

For our action-effect task, the challenge is to generate an output image that both depicts the effect of the action and retains the scene context from the input image. In other words, when peeling the carrot, the kitchen should remain the same before and after.

One way to approach the task would be to treat this as conditional video prediction, extending an input video into the future, as a sequence of new video frames and guided by the provided action.
We explore an alternative approach based on a new generative model.
GLIDE is a recent neural network model  \cite{nichol2021glide} that has two modes of working. In the first, GLIDE generates an image given a piece of text.  In the second, GLIDE inpaints a masked region of an image given a piece of text. This second mode can be used to edit images through delineating regions (masked areas) and describing the new content in natural language.

We use the second mode of operation to undertake the action-effect task. In doing this, there are two critical sub-tasks: (1)
delineate the region in which we expect the effects of the action to be visible; and (2) express the effects of an action as a short textual description. Typically, action datasets provide annotations for actions expressed only as verb-noun pairs, emphasising the action rather than the effect of the action.


 The contributions of our work are as follows:   
\begin{enumerate}[-,nosep]
    \item{Application of the image synthesis model GLIDE to the action-effect task;}
    \item{Consideration of how to select masked regions for inpainting;}
    \item{Consideration of how to map actions into action-effect textual descriptions;}
    \item {Qualitative experiments evaluating the approach on the EPIC dataset.}
    
\end{enumerate}


\section{Background on the action-effect prediction task}


Human action prediction has been a prevalent topic in recent years, with the goal of predicting forthcoming actions from temporally incomplete action videos. There are two primary research directions: predicting the category of a subsequent action and predicting a motion trajectory. Our action-effect prediction is distinct from both of these and can be regarded as a new kind of action prediction task. Here we give a general formulation of the task.

Given the following:
\begin{enumerate}[-,nosep]
    \item An image depicting the initial world state before a human action.
    \item A linguistic description of the action.
\end{enumerate}
Produce an image depicting the final world state following the action.

\begin{figure}[!ht]
\begin{center}
\includegraphics[scale=0.142]{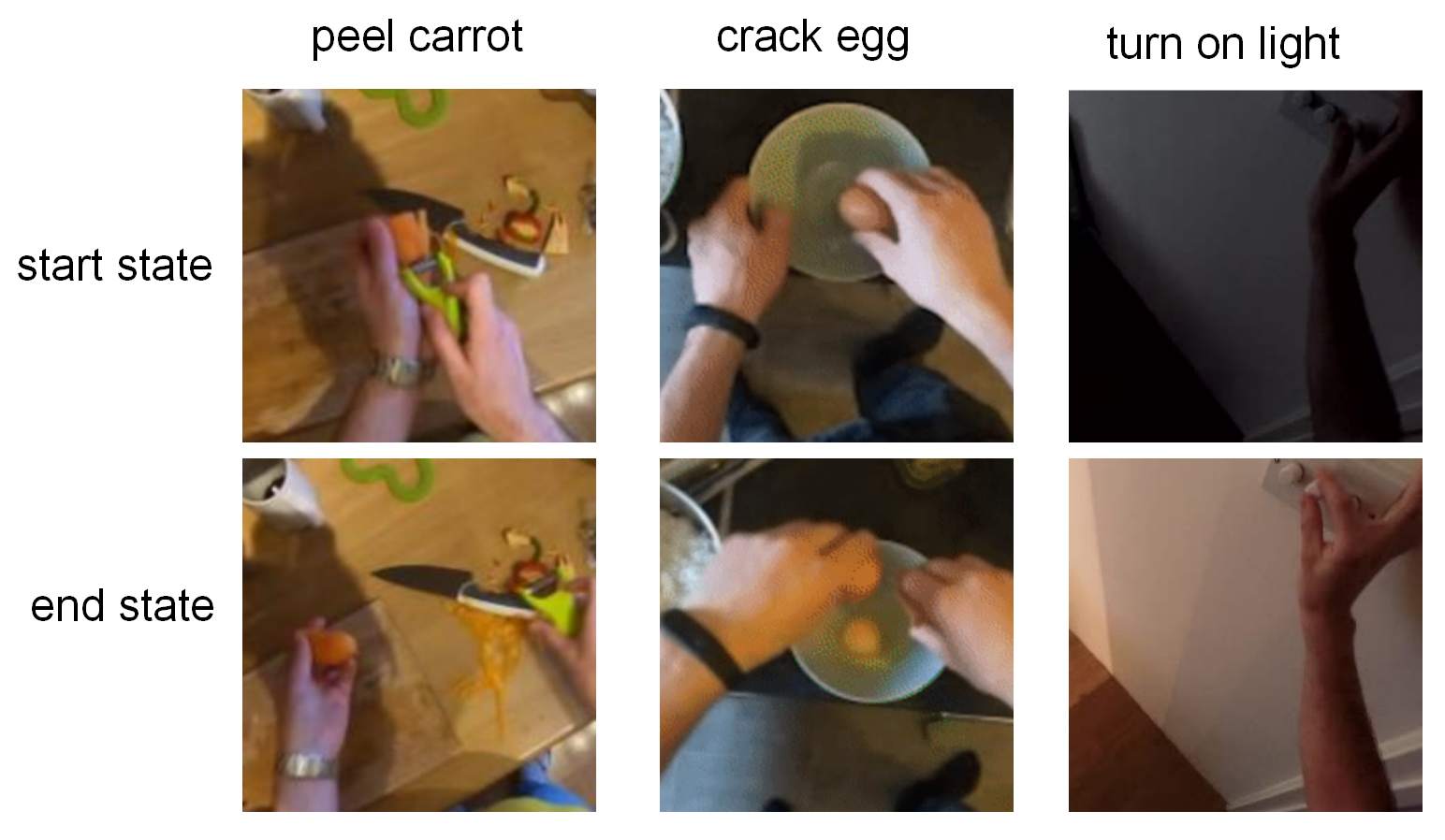} 
\caption{Examples of prediction of the world's future state after an action. The images are taken from the EPIC-Kitchen dataset.}
\label{fig.1}
\end{center}
\end{figure}

For example, in Figure \ref{fig.1}, for the action `crack egg', ``the end result is that the entire contents of the egg will be in the bowl, with the yolk unbroken, and that the two halves of the shell are held in the cook's fingers'' \cite{davis1998naive}. We expect that given a reference image depicting the action's start state and a text prompt about the action `crack egg',  the generative model can predict a future frame depicting the action's effect, that is, the end world state after the action.

Our task can be viewed as a conditional image prediction problem. Thus it may benefit from architectures designed for image synthesis. 
Generative Adversarial Networks (GAN) \cite{goodfellow2014generative} have gained great attention since their introduction in 2014.
Variational auto-encoders (VAE), which were put forward around the same time, 
have also increased in popularity over recent years. 
Recent work on image synthesis using a VAE includes Dall.E \cite{ramesh2021zero}.
Inspired by simulated annealing and diffusion processes, the use of diffusion models in image synthesis \cite{ho2020denoising,dhariwal2021diffusion}
have recently achieved high quality results.

The following logic leads us to focus on the GLIDE model \cite{nichol2021glide}. To begin with, we consider generative models that can take both visual and textual input. Following that, we concentrate on diffusion-based methods \cite{liu2021more} because they have shown superior performance in terms of image sample quality and have well-established model structures that make use of recent advances in transformers and diffusion methods. Finally, we choose GLIDE among diffusion-based models since it is trained on billions of images and can be used to perform image editing (inpainting).



\section{Datasets}

In general, there are two types of human action video datasets: those taken in the third-person and those taken in the first-person (egocentric).


\textbf{Third-person} videos/images human action Datasets include UCF101, KTH, and UCFsports, Human3.6M, Sports1m, Penn Action and THUMOS-15 \cite{zhou2020deep}. These datasets cover human actions like dancing, climbing, walking, etc. The viewpoint is from a third-person standpoint. 

\textbf{First-person (egocentric)} video datasets include  Extended GTEA Gaze+ \cite{li2021eye} and EPIC-Kitchens-100  \cite{damen2020epic}. 
The majority of the actions in these datasets involve first-person observers holding or manipulating objects. 
The actions in these two datasets are all about the preparation of meals in a realistic kitchen scenario. 


We selected egocentric videos for two reasons:
\begin{enumerate}[nosep]
    \item In egocentric videos, most actions are close-ups of hand movements so that the regions of manipulated objects are prominent within the image, which allows for the transmission of sufficient information regarding object state changes after resizing to $64\times64$ as required for the GLIDE model; 
    \item The publicly available version of GLIDE ('filtered') was trained on a filtered version of a dataset that excluded all images of humans, so may have poor performance on whole-body state change prediction. 
\end{enumerate}

EPIC-Kitchens was utilised as the reference dataset in our experiments because the video quality is better (full HD over $1280\times920$ and brighter lightening)  and the dataset covers 100 hours of recording, more than three times the amount of Extended GTEA Gaze+. 

\section{Method}

The proposed method for action-effect prediction using GLIDE depends on two key elements described in the following sections.


\subsection{Setting of Mask Areas}
The success in using GLIDE in the action-prediction task depends critically on the choice of the mask region for inpainting. We consider two alternatives: defining a fixed mask and generating a mask tailored to the content of the given image.

\begin{figure}[!ht]
\begin{center}
\includegraphics[scale=0.132]{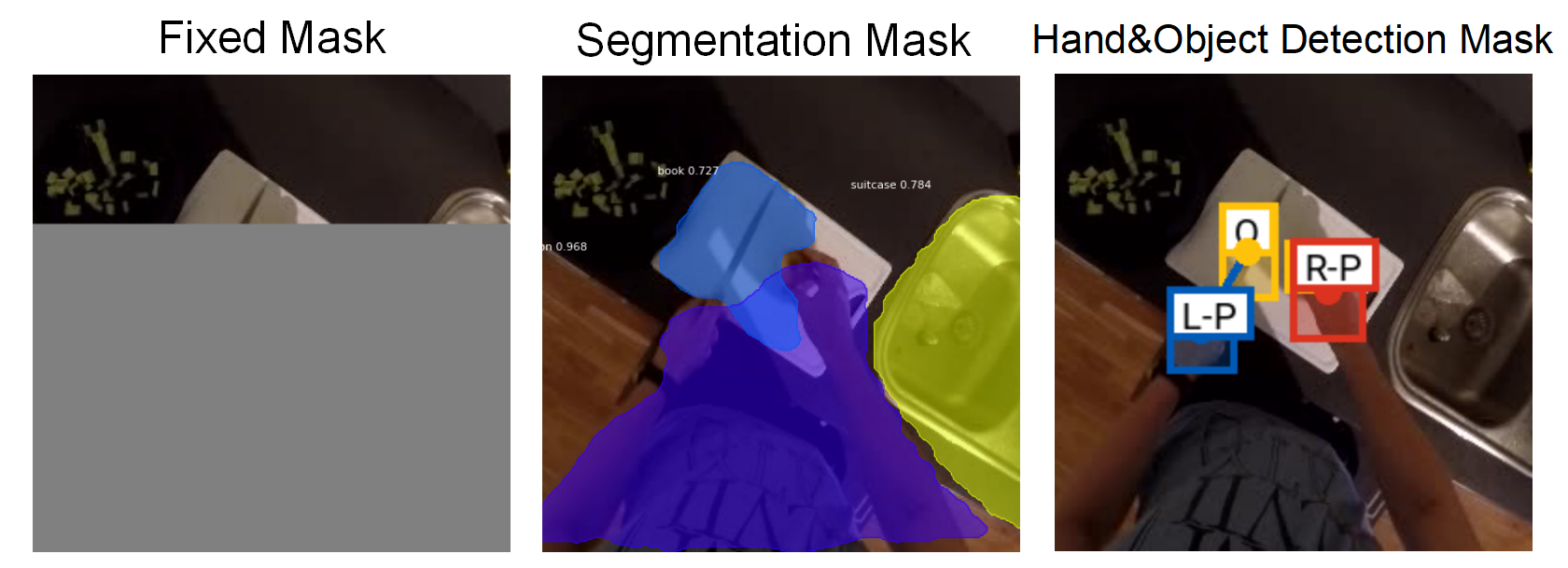} 
\caption{Examples of different mask area settings. }
\label{fig.4}
\end{center}
\end{figure}

\subsubsection{Using a fixed mask}
The direct and easiest way to define a masked region for inpainting is to fix the mask area for all input images. For example, as shown in Figure \ref{fig.4} (left), the mask covers the lower two thirds of the image.

The problem with a fixed mask is that the chosen region may not be appropriate for every instance of an action. If the mask region is too big, it may not include sufficient information about the scene context, and the generated image may not resemble the original scene context, except in the area of the fixed portion. If we set the mask region too small, we cannot be certain that the whole state changes occur in that area.

\subsubsection{Using a generated mask around a region of interest}

For action-effect prediction, the region of interest in an image is the area in which actions are performed. Ideally, we would set the inpainting mask to be this region. The detection of such a region is meaningful because it indicates a zone around the centre of attention, that is where to look for action-relevant items in the scene in order to identify state changes. We adopt two methods for finding masks around regions of interest. In both cases, the regions have already been provided for the EPIC-KITCHENS-100 dataset \footnote{https://github.com/epic-kitchens/epic-kitchens-100-object-masks}  \footnote{https://github.com/epic-kitchens/epic-kitchens-100-hand-object-bboxes} to delineate the prominent objects within the scene.

In the first method, we define object \textbf{segmentation masks} from the regions produced by Mask-RCNN \cite{he2017mask}. 

In the second method, we define \textbf{hand and object masks} from detection boxes around the hands and the manipulated objects using a system \cite{shan2020understanding} based on Faster-RCNN. In our experiments, we filter the detections to accept only those above a significance threshold of 0.1.

\subsection{Generating the text prompt}
The inpainted output image from GLIDE is generated in response to a text prompt, which is a description of the effect of an action. We generate this textual description automatically from the action phrase. To do this, we use the pre-trained auto-regressive language model GPT-3 \cite{brown2020language} to obtain textual descriptions of future world states from action phrases. The input to GPT-3 is a sequence of randomly chosen pairs of action phases with the corresponding textual effect descriptions (two pairs in our experiments), followed by the given action phrase. The continuation of this sequence predicted by GPT-3 provides the textual description we require. We randomly selected the examples from the human collected \cite{gao2018action} action-effect pairs dataset. For example, for the action `cut apple', the generated action effect description is `Apple is cut in half with a knife'.

In experiments, we compare performance with an approach in which the action phrase is input directly to GLIDE as the text prompt.

\section{Results}

We visually compare performance on the action-effect prediction task with the three mask settings and two ways of generating text prompts.

\subsection{Influence of Mask Areas}

\begin{figure*}[!ht]
\begin{center}
\includegraphics[scale=0.153]{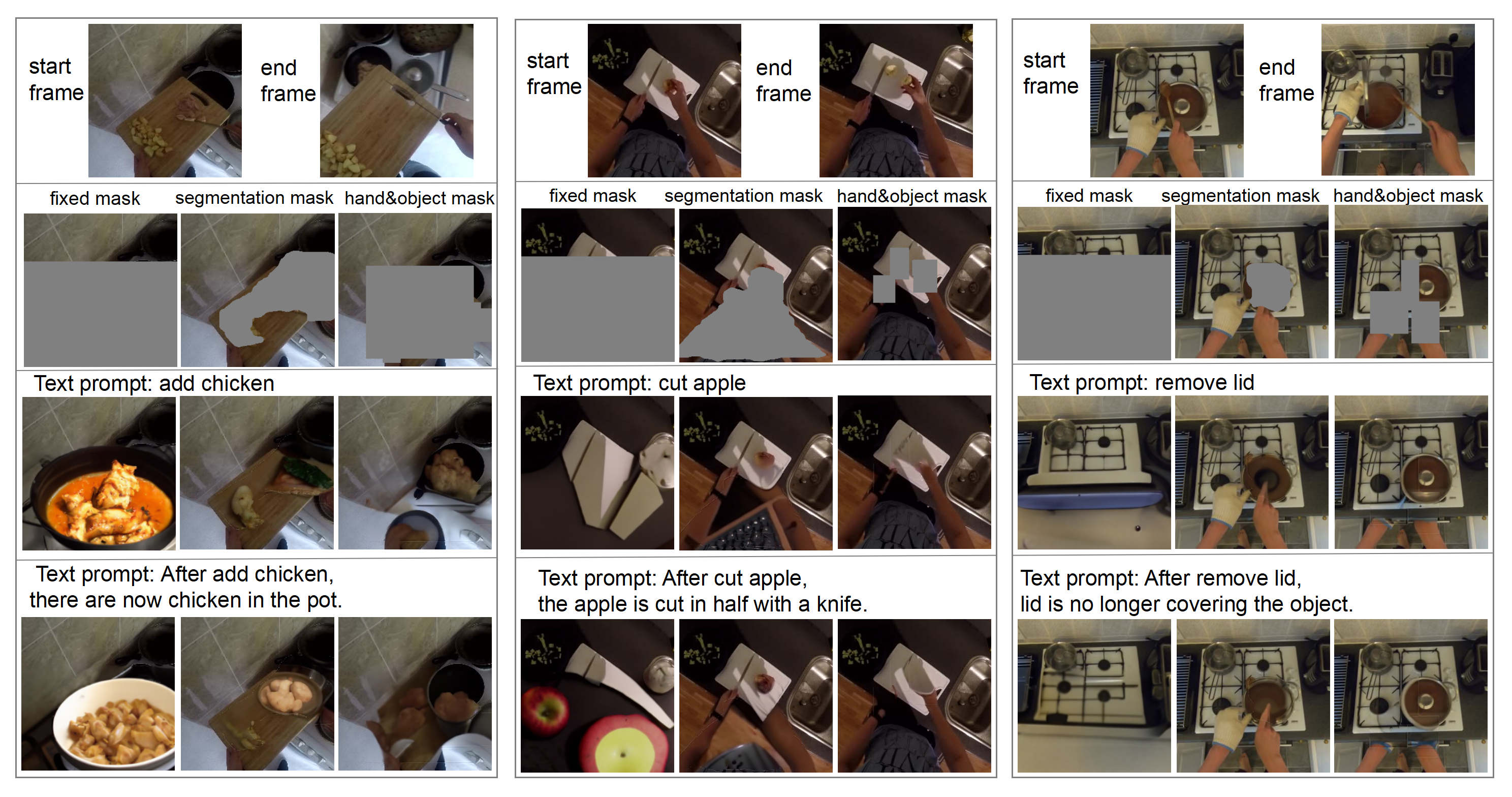} 

\caption{Examples of action-effect prediction on action ``add chicken" (left), ``cut apple" (middle) and ``remove lid" (right) with GLIDE using different masks and text prompts. Within the panel for each action are shown the original start and end frames from the dataset (top row), the three masks (2nd row), the results using the action phrase as the text prompt to GLIDE (3rd row), and the results using the effect description from GPT-3 as the text prompt to GLIDE (4th row)}
\label{fig.10}
\end{center}
\end{figure*}

In Figure \ref{fig.10} we show three different types of action: add, cut, and remove. 
We set the \textbf{fixed mask} to the region that is perceived as the foreground in the majority of action instances. We observe that the GLIDE model with a fixed mask is capable of refilling the masked image with manipulated objects. 
But the generated object, which is `chicken' for `add chicken', `apple' for `cut apple' in Figure \ref{fig.10}, takes the whole unmasked area.

For the \textbf{hand and object masks}, the mask incorporates more information about the environment in comparison to the fixed mask. The objects in action can be projected to have a reasonable size and form. 
However, some vital regions may be cropped owing to the rectangular form of the detection boxes. For the action `remove lid', the object detection area does not fully cover the lid, but rather the movable section.

With \textbf{segmentation masks}, we got better results on these three action instances. 
For action `add chicken', apart from the manipulated object (chicken), potato and pot are also masked. The masks are more precise, and there is more visual information: part of hand, the chopping board and kitchen environment, allowing it to refill the pot and chopping board. The resulting picture is more compatible with its environment.
For action `cut apple', the apple is predicted to be of a suitable size and location, but the hand is not created in a sensible way.
For action `remove lid', the pot is well detected compared with using fixed and detection masks. Though the pot shape isn't quite round and the borders aren't perfectly connected, it best describes the lid removed state.

While mask design improves prediction, there is still room for improvement: the model cannot include any information about the manipulated object, thus the newly produced objects are not exactly those that appeared in the start frame.


\subsection{Influence of Text Prompts}

The effect description for the action ``add chicken" as shown in Figure \ref{fig.10} comes from GPT-3.
In comparison to a pure action phrase, the text prompt ``After add chicken, there are now chicken in the pot." contains more detailed information regarding the effects of an action, specifically that the  chicken is now in the pot. We can observe that, with this text prompt, in all predicted images, the chicken is in the pot. We can also observe apparent improvement in generation results with a fixed mask on the action ``cut apple" and segmentation mask on action ``remove lid".
We can see a noticeable improvement in generation image quality on action ``cut apple" with fixed mask and action ``remove lid" with segmentation mask.

\subsection{Failure cases}

\begin{figure}[!ht]
\begin{center}
\includegraphics[scale=0.12]{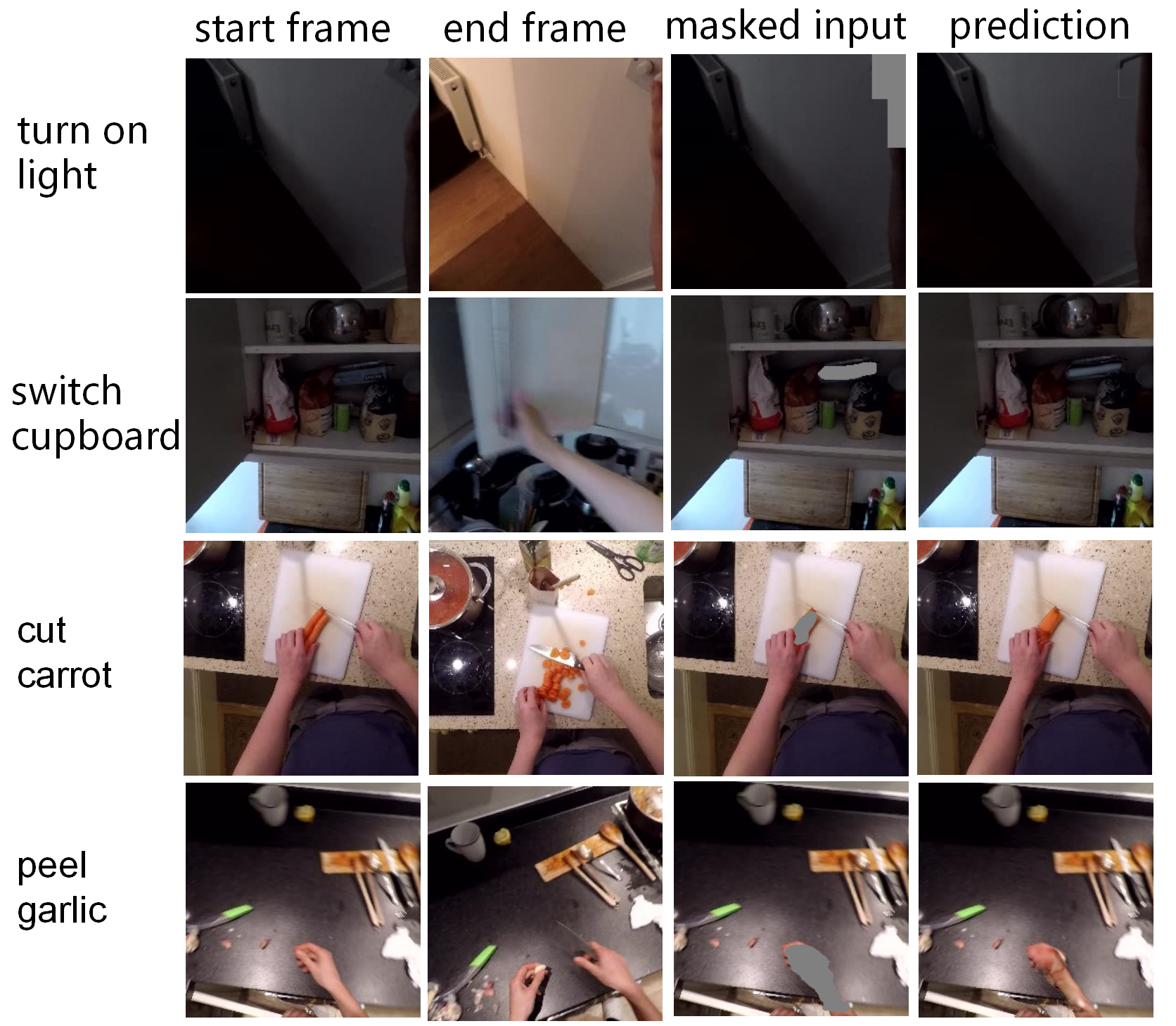} 
\caption{\fj{Failure examples using segmentation mask and action phrase as text prompt.} }
\label{fig.11}
\end{center}
\end{figure}

\fj{In Figure \ref{fig.11} we show several failure cases: some actions that involve changing the brightness of the environment rather than changing the attributes of items, e.g., `turn on light'; certain position-changing actions such as `switch cupboard' (i.e. open or close cupboard); and object-quantity-increasing actions such as `cut carrots' and `peel garlic', the initial masked area may be insufficiently large to fully fill in the newly formed pieces.}

\section{Conclusions and Future Work}
We have explored GLIDE's potential on our real-world action-effect prediction task. We have shown that by optimising the mask area design and converting actions into action-effect descriptions as text prompts, the GLIDE model can create more accurate predictions that are consistent with the start world state.

In future work, we plan to fine-tune GLIDE for our action-effect task using a specialised dataset. It would also be interesting to explore whether GLIDE could be developed to avoid the use of a mask and instead revise the whole image based on a text prompt.

\section{References}


%



%
%
%
%
%

\bibliographystyle{lrec2022-bib}
\bibliography{lrec2022-example}

\end{document}